# Found-RL: foundation model-enhanced reinforcement learning for autonomous driving

Yansong Qu[a], Zihao Sheng[b], Zilin Huang[b], Jiancong Chen[a], Yuhao Luo[b], Tianyi Wang[c], Yiheng Feng[a], Samuel Labi[a], Sikai Chen[b*]

[a] *Lyles School of Civil and Construction Engineering, Purdue University, West Lafayette, 47907, USA*

[b] *Department of Civil and Environmental Engineering, University of Wisconsin-Madison, Madison, 53706, USA*

[c] *Department of Civil, Architectural, and Environmental Engineering, University of Texas at Austin, Austin, 78712, USA*

\* Corresponding author.

*E-mail address: sikai.chen@wisc.edu*

**Abstract**

Reinforcement Learning (RL) has emerged as a dominant paradigm for end-to-end autonomous driving (AD) with real-time inference. However, RL typically suffers from sample inefficiency and a lack of semantic interpretability in complex scenarios. To mitigate these limitations, Foundation Models (particularly, Vision-Language Models (VLMs)) can be integrated because they offer rich, context-aware knowledge. Yet still, deploying such computationally intensive models within high-frequency multi-environment RL training loops is severely hindered by prohibitive inference latency and the absence of unified integration platforms. To bridge this gap, we present Found-RL, a specialized platform tailored to leverage foundation models to efficiently enhance RL for AD. A core innovation of the proposed platform is its asynchronous batch inference framework, which decouples heavy VLM reasoning from the simulation loop. This design effectively resolves latency bottlenecks, supporting real-time or near-real-time RL learning from VLM feedback. Using the proposed platform, we introduce diverse supervision mechanisms to address domain-specific challenges: we first implement Value-Margin Regularization (VMR) and Advantage-Weighted Action Guidance (AWAG) to effectively distill expert-like VLM action suggestions into the RL policy. Furthermore, for dense supervision, we adopt high-throughput CLIP for reward shaping. We mitigate CLIP's dynamic blindness and probability dilution via Conditional Contrastive Action Alignment, which conditions prompts on discretized speed/command and yields a normalized, margin-based bonus from context-specific action-anchor scoring. Found-RL delivers an end-to-end pipeline for fine-tuned VLM integration with modular support, and shows that a lightweight RL model with millions of parameters can achieve near-VLM performance compared with billion-parameter VLMs while sustaining real-time inference (~500 FPS). Code, data, and models will be publicly available at https://github.com/ys-qu/found-rl.

**Keywords**

**Introduction**

The pursuit of robust autonomous driving requires agents that can navigate complex, dynamic environments with human-like understanding (Qu et al., 2025b). Traditional rule-based approaches struggle with corner cases. On the other hand, data-driven approaches, particularly RL, have emerged as a promising paradigm for learning end-to-end control policies through interaction with environments. However, training RL agents from the ground up is akin to a student attempting "self-study" with a textbook (Lake et al., 2017; Sutton and Barto, 1998); the process is often inefficient and "scattergun" (Fig. 1 (a)), relying on exhaustive trial-and-error. To mitigate this, Human-in-the-loop RL (Huang et al., 2024a, 2025a; Peng et al., 2024) introduces human experts to correct behaviors, acting as a private tutor. While this "Ask your teacher" approach is accurate (Fig. 1 (b)), it is intrinsically demanding and unscalable due to the high cost of human labor and attention fatigue.

The recent surge in the use of Foundation Models, specifically VLMs, offers a transformative third path: Foundation Model-enhanced RL. Foundation Models are large, pre-trained models trained on massive and diverse datasets and can be adapted to many downstream tasks. As illustrated in Fig. 1 (c), these models function as a "Tireless Mentor" because they possess the semantic reasoning capabilities of a human teacher while remaining "Always Ready" to provide scalable feedback. VLMs can interpret complex driving scenes and generate meaningful supervision signals, potentially combining the data efficiency of imitation learning with the self-exploration of RL. Despite this promise, integrating these computationally heavy models into multi-environment high-frequency RL pipelines remains a significant engineering hurdle. The inference latency and the lack of unified frameworks have largely confined such research to theoretical exploration rather than efficient, closed-loop training.

To bridge this gap, we propose Found-RL, a platform tailored to foundation model-enhanced RL in autonomous driving. Found-RL provides a streamlined pipeline spanning from CARLA (Dosovitskiy et al., 2017)-based simulation to specialized training modules. Unlike generic implementations, the proposed platform features a novel asynchronous batch inference framework which decouples heavy VLM reasoning from the real-time simulation loop. This design effectively resolves the latency bottleneck, enabling real-time or near-real-time training with online supervisions from VLMs.

Found-RL also serves as a versatile testbed for diverse supervision mechanisms, we systematically explored multiple strategies to leverage VLM feedback, specifically implementing VLM-action guidance to provide online expert actions and VLM-based reward shaping to densify sparse environmental rewards with semantic understanding. By unifying efficient engineering with flexible algorithm support, Found-RL lowers the barrier for researchers to validate new architectures, ultimately advancing the development of the next generation of semantically aware autonomous driving agents.

The main contributions of this work are summarized as follows:

**(1) A Unified Platform for Foundation Model-Enhanced RL**: We propose Found-RL, a framework specifically tailored for autonomous driving. It features a novel asynchronous batch inference architecture that decouples computationally intensive VLM reasoning from the RL simulation loop, successfully resolving latency bottlenecks and enabling efficient, real-time or near-real-time closed-loop RL training with VLMs.

**(2) VLM Action Guidance Mechanisms**: To effectively leverage expert-like action advice from VLMs, we introduce Value-Margin Regularization (VMR) and Advantage-Weighted Action Guidance (AWAG). These mechanisms facilitate the distillation of VLMs' action knowledge into the RL policy, significantly improving exploration efficiency and decision-making quality.

**(3) CLIP-based Reward Shaping**: We introduce Conditional Contrastive Action Alignment, which conditions CLIP (Radford et al., 2021) text prompts on discretized ego speed and route command and scores a small context-specific set of action anchors to derive a normalized, margin-based reward bonus, mitigating CLIP's dynamic blindness and probability dilution while providing dense supervision.

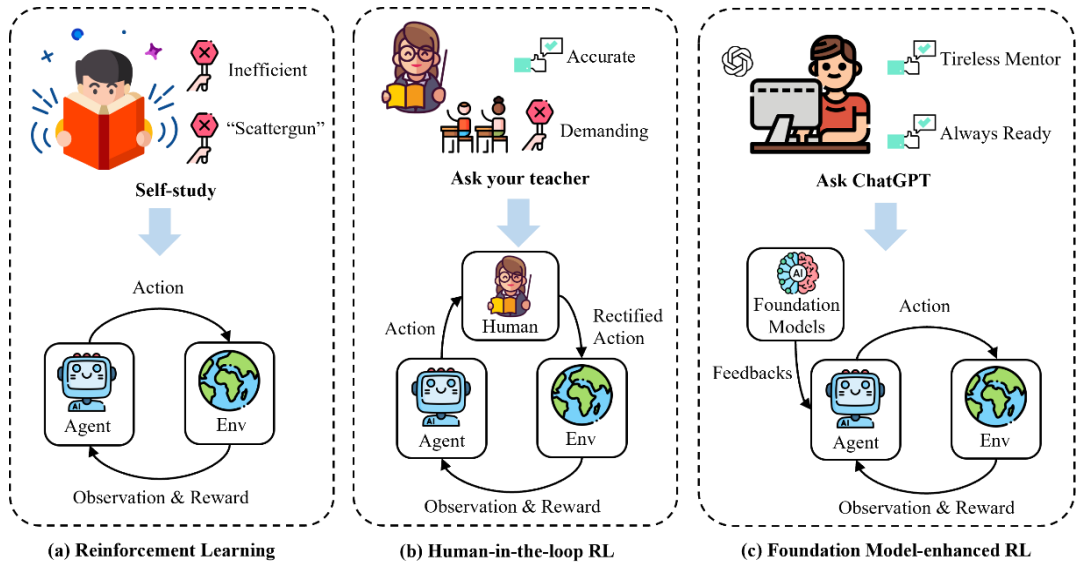

**Fig. 1.** Conceptual comparison of Reinforcement Learning, Human-in-the-loop RL, and Foundation Model-enhanced RL.

## Related work

In this section, we review three related threads: reinforcement learning-based autonomous driving, foundation models in autonomous driving, and simulation platforms for autonomous driving. We focus on these aspects because our work lies at their intersection: RL provides the learning framework, foundation models offer semantic priors to guide training, and simulation platforms enable scalable and reproducible evaluation.

*2.1. Reinforcement learning-based autonomous driving*

RL has emerged as a promising paradigm for autonomous driving due to its ability to learn complex sequential decision-making policies through interaction with the environment. Its potential to optimize long-term objectives without requiring explicit supervision makes it well-suited for handling dynamic and uncertain traffic scenarios (Qu et al., 2025a).

Numerous studies (Sheng et al., 2025b, 2025c) have applied RL to various aspects of autonomous driving, including decision-making, scenario generation, etc. However, most existing works (Sheng et al., 2024) are developed in isolated environments with customized pipelines, leading to limited reproducibility and comparability. Moreover, despite the recent success of foundation models such as VLMs and Large-Language Models (LLMs), mainstream RL approaches (Long et al., 2024) still heavily rely on self-exploration and sparse reward feedback. They often overlook the potential of expert-guided learning (Peng et al., 2022) or semantically-rich reward shaping mechanisms (Huang et al., 2024b) to accelerate and stabilize policy learning, especially in safety-critical domains.

To address these gaps, we propose a VLM-enhanced platform that supports training, evaluation, and benchmarking of RL-based autonomous driving agents. Our platform enables plug-and-play integration of VLMs to guide exploration and augment reward design, significantly improving the performance of traditional RL algorithms such as SAC (Haarnoja et al., 2018) and DrQv2 (Yarats et al., 2021). Extensive experiments across diverse benchmarks demonstrate the effectiveness and generalizability of our framework.

*2.2. Foundation models in autonomous driving*

Foundation models are pre-trained on large-scale and diverse data, and have shown strong generalization ability across domains. When fine-tuned with domain-specific data, they can be adapted into specialists that perform well in targeted tasks. Recent efforts such as DeepSeek-R1 (Guo et al., 2025) have focused on using RL to enhance the reasoning and generalization capabilities of LLMs. In contrast, our focus is on the reverse direction, leveraging foundation models to assist RL agents.

In the context of autonomous driving, VLMs and LLMs have been applied to tasks such as semantic understanding (Sheng et al., 2025a), language-guided decision-making (Jiang et al., 2024), and natural language explanations of driving behavior (Xu et al., 2024). For example, VLMs can match visual inputs from the driving environment with natural language goals or instructions, helping the agent to navigate complex scenarios in a more human-aligned manner. LLMs have been used to interpret and explain driving decisions, define high-level goals, and assess the plausibility of planned trajectories (Zhang et al., 2024). These applications highlight the potential of VLMs and LLMs to improve transparency, flexibility, and safety in autonomous driving systems.

To bridge the gap between foundation models and RL, we present Found-RL, a platform designed to integrate VLMs into RL pipelines for autonomous driving. Found-RL enables training, evaluation, and benchmarking with support for plug-and-play modules based on foundation models. These modules can be used to guide exploration, provide reward signals and so on. By enabling broad integration of foundation models, Found-RL serves as a flexible and extensible platform that supports the autonomous driving community in advancing research at the intersection of RL and foundation model technologies.

*2.3. Simulation platforms for autonomous driving*

Several frameworks have been developed to support RL research, such as Stable-Baselines3 (SB3) (Raffin et al., 2021), RLlib (Liang et al., 2018), and CleanRL (Huang et al., 2022). These libraries offer modular and well-tested implementations of popular RL algorithms, making them accessible for general-purpose experimentation. However, they are not specifically designed for autonomous driving scenarios and often lack support for the complex, high-dimensional, and safety-critical environments that characterize this domain.

To address this gap, platforms such as CARLA Leaderboard (Dosovitskiy et al., 2017) and Sky-Drive (Huang et al., 2025b) provide simulation-based environments and evaluation protocols tailored to autonomous driving. These platforms allow for policy training, testing, and benchmarking under realistic driving scenarios. Nevertheless, they have yet to integrate emerging foundation model technologies, such as VLMs, which have shown strong potential in improving sample efficiency, semantic understanding, and planning capabilities. To enable deeper integration of foundation models into RL workflows for autonomous driving, we introduce a new platform Found-RL.

Our platform is designed to support foundation model-enhanced RL with features that address current bottlenecks in model-agent interaction. Specifically, we implement asynchronous batch inference, which allows VLMs to process observations in parallel with environment simulation, ensuring that the environment is not paused while waiting for model outputs. This design effectively mitigates the inference-induced delays that typically hinder real-time training. The platform supports CLIP (Radford et al., 2021)-based reward computation, expert action guidance from VLMs, and can be extended to support more applications. Together, these capabilities establish a flexible and extensible foundation for advancing RL in autonomous driving with integrated foundation model support.

**Preliminaries**

In this section, we first introduce the problem formulation of foundation model-enhanced RL, and we then illustrate the off-policy actor-critic learning. These serve as the base for the explanation of our proposed method.

*3.1. Problem Formulation*

We formulate an autonomous driving task as a discounted Markov decision process $\mathcal{M} = (\mathcal{O}, \mathcal{A}, P, r, \gamma)$, where $o_t \in \mathcal{O}$ is the observation at time $t$, $a_t \in \mathcal{A}$ is the action, $r_t = r(o_t, a_t)$ is the environment reward, and $\gamma \in (0,1)$ is the discount factor. A policy $\pi_\phi(a \mid o)$ induces a trajectory distribution through the transition dynamics $P(o_{t+1} \mid o_t, a_t)$. The goal is to learn a policy that maximizes the expected discounted return

$$J(\pi_\phi) = \mathbb{E}_{\pi_\phi}\left[\sum_{t=0}^{\infty} \gamma^t r_t\right]. \tag{1}$$

In VLM-enhanced RL, a VLM provides auxiliary feedback signals derived from the current observation and context. We denote the feedback at time $t$ as $g_t = \mathcal{F}_\psi(o_{\leq t})$, which can represent an action suggestion or action prior, a critique signal, or a constraint-related indicator. This yields an augmented formulation where the RL agent conditions on $\tilde{o}_t = (o_t, g_t)$, and optimization can incorporate the VLM guidance either through reward shaping $\tilde{r}_t = r_t + \lambda r^{\text{vlm}}(g_t)$ or through policy regularization toward a VLM action prior $\pi_{\text{vlm}}(a \mid o_t, g_t)$. A generic objective that captures these uses is

$$\max_\phi \mathbb{E}\left[\sum_{t=0}^{\infty} \gamma^t \left(r_t + \lambda r^{\text{vlm}}(g_t)\right)\right] - \beta \mathbb{E}\left[\text{KL}\left(\pi_\phi(\cdot \mid \tilde{o}_t) \| \pi_{\text{vlm}}(\cdot \mid \tilde{o}_t)\right)\right], \tag{2}$$

where $\lambda$ weights the VLM-based reward shaping term and $\beta$ controls the KL regularization toward the VLM action prior.

### 3.2. Off-policy actor-critic learning

We consider a discounted Markov decision process $\mathcal{M} = (\mathcal{O}, \mathcal{A}, P, r, \gamma)$ and a parameterized stochastic policy $\pi_\phi(a \mid o)$. Off-policy actor-critic methods learn an action-value function $Q_\theta(o, a)$ and optimize $\pi_\phi$ using transitions stored in a replay buffer $D$, where $D$ contains tuples $(o_t, a_t, r_t, o_{t+1}, d_t)$ collected by a behavior policy that can differ from $\pi_{\phi_1}$ and $d_t \in \{0,1\}$ indicates episode termination. The critic is learned via temporal-difference bootstrapping with a slowly updated target network $Q_{\bar{\theta}}$, using the one-step Bellman target:

$$y_t = r_t + (1 - d_t)\gamma \mathbb{E}_{a' \sim \pi_\phi(\cdot \mid o_{t+1})}[Q_{\bar{\theta}}(o_{t+1}, a')], \tag{3}$$

and minimizing the squared Bellman error:

$$L_\theta = \mathbb{E}_{(o_t, a_t, r_t, o_{t+1}, d_t) \sim D}(Q_\theta(o_t, a_t) - y_t)^2. \tag{4}$$

The actor is updated to maximize the critic value of its actions for states sampled from the replay buffer, which can be written as minimizing

$$L_\phi = -\mathbb{E}_{o_t \sim D}\left[\mathbb{E}_{a \sim \pi_\phi(\cdot \mid o_t)}[Q_\theta(o_t, a)]\right]. \tag{5}$$

In practice, off-policy actor-critic algorithms commonly incorporate additional stabilizers such as clipped double-Q learning, target-policy smoothing, and entropy regularization, while retaining the same core structure of learning $Q_\theta$ from replay and improving $\pi_\phi$ using the critic.

**Method**

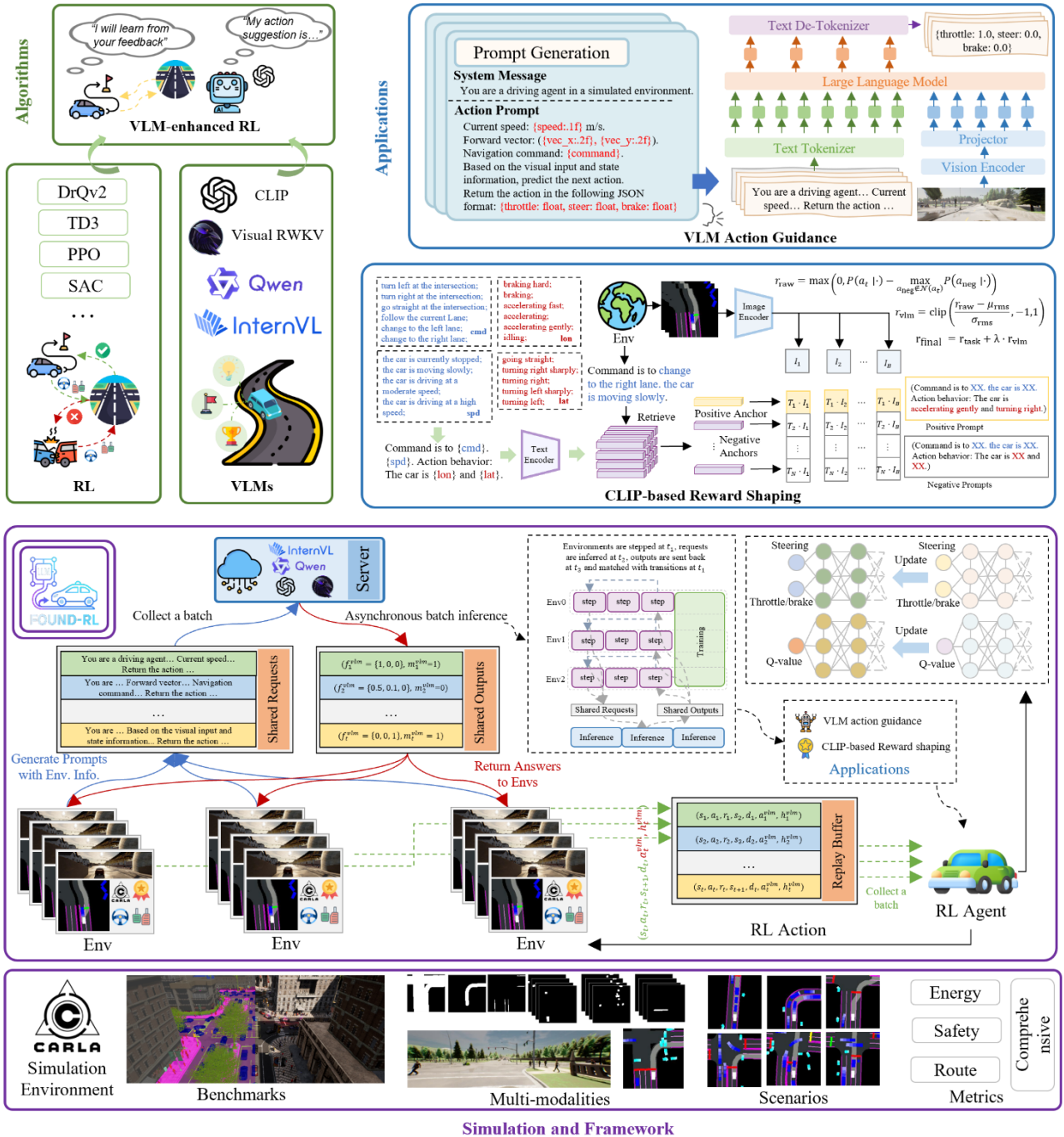

**Fig. 2.** Overall framework.

Fig. 2 summarizes the overall framework of Found-RL as a unified platform with three coupled parts (simulation, algorithms, and applications), designed to make foundation model-enhanced RL practical for autonomous driving.

In the simulation part, Found-RL builds on CARLA to provide standardized benchmarks and diverse traffic scenarios, producing multi-modal driving observations (e.g., bird's-eye view (BEV) or BEV masks) together with lightweight structured context from the simulator (such as speed and traffic-rule states) and task-defined rewards/termination signals. In the algorithms part, the platform integrates classical RL learners with foundation models via an asynchronous batch inference pipeline: rollout workers turn observations and context into prompts, send them to a shared request queue, and an inference server micro-batches requests across parallel environments to run VLM/CLIP inference and return feedback without blocking the simulator loop; the resulting feedback (e.g., expert-like action suggestions or semantic scores) is stored in the replay buffer alongside standard transitions and consumed by modular learning objectives. In the applications part, Found-RL instantiates these modules into end-to-end workflows for learning from foundation-model feedback, including VLM action guidance and CLIP-based reward shaping, and supports systematic benchmarking/ablation under consistent evaluation protocols with comprehensive route, safety, and efficiency metrics.

## 4.1. Asynchronous batch inference framework

VLM inference is substantially more expensive than a single simulator step or an actor-critic forward pass, so issuing a separate VLM query for every environment step quickly becomes the dominant training bottleneck. To make VLM feedback practical at scale, we implement an asynchronous batch inference framework that decouples environment rollout from VLM inference through a client-server design, and batches requests across parallel environments and across time, as illustrated in Fig. 2. At each step $t$ in environment $e$, the rollout worker constructs a query prompt $p_t^e$ from the current observation and lightweight environment metadata, and sends it to a shared request queue

$$p_t^e = \mathcal{P}(o_t^e, \eta_t^e), q_t^e = (e, t, p_t^e), q_t^e \to \mathcal{Q}_{req}, \tag{6}$$

where $\eta_t^e$ can include speed, route command, traffic-light state, and other structured signals already available from the simulator. An inference server continuously pops requests from $Q_{\text{req}}$ and forms micro-batches using a size cap $B_{\max}$ and a short timeout $\tau$, which prevents stragglers from stalling the pipeline. $\mathcal{B}_k$ can be defined as follows:

$$\mathcal{B}_k = \text{Batch}(Q_{\text{req}}; B_{\max}, \tau) = \{q_{k,1}, \dots, q_{k,B_k}\}. \tag{7}$$

Given a batch $\mathcal{B}_k$, the server runs the selected VLM $\mathcal{F}_\psi$ in parallel over the batch and returns the per-request outputs through a shared output queue $Q_{\text{out}}$. In our setting, the VLM output is represented as an VLM feedback $f_t^{\text{vlm},e}$ (e.g., VLM expert action $a_t^{\text{vlm},e}$) with an availability indicator $m_t^e \in \{0,1\}$, which supports missing or delayed responses without breaking training. $(f_t^{\text{vlm},e}, m_t^e)$ can be defined as follows:

$$(f_t^{\text{vlm},e}, m_t^e) = \mathcal{F}_\psi(p_t^e), (e, t, a_t^{\text{vlm},e}, m_t^e) \to Q_{\text{out}}. \tag{8}$$

On the rollout side, outputs are asynchronously matched back to environments through the key $(e,t)$. The resulting transition stored in the replay buffer is augmented with the VLM fields:

$$(o_t, a_t, r_t, o_{t+1}, d_t, f_t^{\text{vlm}}, m_t) \in D, \tag{9}$$

This way, the downstream learning components can consume VLM guidance through masked objectives whenever $m_t = 1$. This design makes VLM inference an overlapped service rather than a blocking per-step call, and batching amortizes the fixed overhead of tokenization, vision encoding, and decoding across several requests.

In practice, this framework allows us to scale VLM feedback to many parallel environments while keeping the actor-critic training loop fully utilized and robust to variable VLM latency.

### *4.2. VLM action guidance*

We introduce VLM action guidance as a form of constrained exploration, where a VLM provides step-wise action proposals to restrict the policy's search space and improve early-stage learning efficiency. In our implementation, we use a fine-tuned InternVL3-1B (Zhu et al., 2025) (see Section 5.2 Vision-Language Models for details) as the default guidance model for its quality-latency trade-off, while the guidance interface is model-agnostic and can be replaced by other instruction-following VLMs with the same input–output format. Found-RL currently supports Value-Margin Regularization (VMR) and Advantage-weighting Action Guidance (AWAG).

*(1) Value-Margin Regularization*

We augment the standard off-policy actor-critic critic update with a value-margin regularizer by sampling extended transitions $(o_t, a_t, r_t, o_{t+1}, d_t, a_t^{\text{vlm}}, m_t) \sim D$, where $a_t^{\text{vlm}}$ is the VLM-proposed feedback $f_t^{\text{vlm}}$ stored for the same state $o_t$ and $m_t \in \{0,1\}$ indicates whether this VLM action is available. For the Temporal Difference (TD) target, we sample the next action from the current actor with scheduled exploration noise, and compute the clipped double-Q bootstrap:

$$y_t = r_t + (1 - d_t)\gamma \min_{i \in \{1,2\}} Q_{\bar{\theta}_i}(o_{t+1}, \tilde{a}_{t+1}), \tilde{a}_{t+1} \sim \pi_\phi(\cdot \mid o_{t+1}). \tag{10}$$

The critic is trained with the usual TD regression:

$$L_{\text{TD}} = \sum_{i \in \{1,2\}} \frac{1}{2} \mathbb{E}_D \left[ \left(Q_{\theta_i}(o_t, a_t) - y_t\right)^2 \right], \tag{11}$$

and when $m_t = 1$ we add a VMR term that encourages the critic to assign a higher value to the VLM action than to the replay action by a fixed margin $\Delta > 0$,

$$L_{\text{VMR}} = \sum_{i \in \{1,2\}} \mathbb{E}_D \left[ m_t \left( Q_{\theta_i}(o_t, a_t^{\text{vlm}}) - \left(\text{stopgrad}\left(Q_{\theta_i}(o_t, a_t)\right) + \Delta\right) \right)^2 \right] \tag{12}$$

The final critic objective is:

$$L_{\text{critic}} = L_{\text{TD}} + \lambda(\text{step}) L_{\text{VMR}} \tag{13}$$

where $\text{stopgrad}(\cdot)$ detaches the replay-action value so the auxiliary term primarily pushes up $Q(o_t, a_t^{\text{vlm}})$. We apply a cosine-decayed coefficient for $\lambda(\text{step})$, with $p(\text{step}) = \min(\text{step}/T, 1)$, $\tilde{p}(\text{step}) = p(\text{step})^\kappa$, and

$$\lambda(\text{step}) = \lambda_{\text{end}} + (\lambda_{\text{start}} - \lambda_{\text{end}}) \cdot \frac{1 + \cos(\pi \tilde{p}(\text{step}))}{2} \tag{14}$$

The decay is necessary in practice because $L_{\text{VMR}}$ can continuously push Q-values upward and increasingly causes bias of the policy toward the VLM action. This reduces effective action exploration and could impair long-horizon learning if the regularizer remains strong throughout the training phase.

*(2) Advantage-weighting Action Guidance*

Similar to VMR, we sample extended transitions:

$$(o_t, a_t, r_t, o_{t+1}, d_t, a_t^{\text{vlm}}, m_t) \sim D, \tag{15}$$

where $a_t^{\text{vlm}}$ denotes the VLM-proposed expert action aligned with $o_t$, and $m_t \in \{0,1\}$ indicates whether this transition contains a valid VLM action.

At update step, we sample $\hat{a}_t \sim \pi_\phi(\cdot \mid o_t)$ and evaluate the clipped double critic:

$$Q(o_t, \hat{a}_t) = \min_{i \in \{1,2\}} Q_{\theta_i}(o_t, \hat{a}_t). \tag{16}$$

When AWAG is enabled, we use an adaptive scale factor, $\lambda$, computed from the first critic head and detached from gradients,

$$\lambda = \frac{\alpha}{\text{stopgrad}\left(\mathbb{E}_{(o_t,a_t) \sim \mathcal{D}}\left[\|Q_{\theta_1}(o_t, a_t)\|_2\right]\right) + \epsilon}, \tag{17}$$

leading to the base actor loss expressed as follows:

$$\mathcal{L}_{\text{base}}(\phi) = -\mathbb{E}_D[\lambda \cdot Q(o_t, \hat{a}_t)]. \tag{18}$$

To incorporate VLM action guidance, we only use samples with $m_t = 1$. Let $a_t^\pi$ be the actor forward output (deterministic action used for advantage estimation). We compute $q_\pi$ as:

$$q_\pi = \min_{i \in \{1,2\}} Q_{\theta_i}(o_t, a_t^\pi), \quad q_{\text{vlm}} = \min_{i \in \{1,2\}} Q_{\theta_i}(o_t, a_t^{\text{vlm}}), \tag{19}$$

and define the advantage, $A_t$, as follows:

$$A_t = q_{\text{vlm}} - q_\pi. \tag{20}$$

We apply a Q-filter gate $g_t = \mathbf{1}\{A_t > 0\}$ so that the policy imitates VLM actions only when they are estimated to outperform the current policy action under the critic. The per-sample weight, $w_t$, is computed as:

$$w_t = \text{clip}\left(\exp\left(\frac{A_t}{\beta}\right), w_{\max}\right), \tag{21}$$

with $\beta = 2$ and $w_{\max} = 20$. In accordance with the implementation, $A_t$ and $w_t$ are treated as constants via stopgrad($\cdot$). The guidance loss maximizes the likelihood of $a_t^{\text{vlm}}$ under the current stochastic policy:

$$\mathcal{L}_{\text{AWAC}}(\phi) = -\mathbb{E}_D\left[m_t \cdot g_t \cdot \text{stopgrad}(w_t) \cdot \log \pi_\phi\left(a_t^{\text{vlm}} \mid o_t\right)\right]. \tag{22}$$

Finally, the actor is updated by combining the base RL term and the VLM guidance term,

$$\mathcal{L}_{\text{actor}}(\phi) = \mathcal{L}_{\text{base}}(\phi) + \iota(\text{step})\mathcal{L}_{\text{AWAC}}(\phi), \tag{23}$$

where $\iota(\text{step})$ follows the cosine-decayed schedule used in our implementation:

$$\begin{aligned} p(\text{step}) &= \min\left(\frac{\text{step}}{T}, 1\right), \tilde{p}(\text{step}) = p(\text{step})^\gamma \\ \iota(\text{step}) &= \iota_{\text{end}} + (\iota_{\text{start}} - \iota_{\text{end}}) \cdot \frac{1 + \cos(\pi \tilde{p}(\text{step}))}{2} \end{aligned} \tag{24}$$

with a faster decay in the early stage.

### *4.3. VLM-based reward shaping*

Leveraging VLM-based reward shaping has emerged as a promising paradigm to provide dense, semantically meaningful supervision (Huang et al., 2024b). Within this domain, CLIP-based approaches are particularly favored for their high inference throughput. However, directly applying CLIP models to continuous control tasks remains challenging due to their inherent limitations in perceiving dynamic states in the form of continuous numerical values (e.g., distinguishing a throttle of 0.99 from 1.0). To address these limitations of CLIPs (Radford et al., 2021), we introduce a **Conditional Contrastive Action-Alignment Reward**. We discretize the vehicle's speed into 4 categories and navigation goals into 6 commands, defining the context at timestep $t$ as $c_t = (\text{cmd}_t, \text{spd}_t)$. To explicitly compensate for the vision encoder's inability to perceive ego-states, we combine every possible context tuple with a compact semantic action space $\mathcal{A}_{CLIP}$ of 6 longitudinal actions (e.g., "idling", "braking hard") and 5 lateral actions (e.g., "turning left sharply"), which result in a comprehensive library of $6 \times 4 \times 6 \times 5 = 720$ context-aware prompts. At each timestep, rather than performing classification over the entire global space, which would lead to severe probability dilution, we dynamically retrieve only the specific subset of $6 \times 5 = 30$ action anchors aligned with the current $c_t$. This conditional slicing mechanism ensures that the text embeddings effectively encode the necessary state information invisible to the image encoder, while simultaneously maintaining a concentrated and discriminative probability distribution over the relevant action candidates. See Fig. 2 (CLIP-based reward shaping) for visual demonstration and more details can be found in Appendix A.

The probability of a specific action $a \in \mathcal{A}_{CLIP}$ given the visual observation $I_t$ is computed as:

$$P(a \mid I_t, c_t) = \frac{\exp\left(\tau \cdot \cos\left(E_I(I_t), E_T(T_{c_t,a})\right)\right)}{\sum_{a' \in \mathcal{A}_{CLIP}} \exp\left(\tau \cdot \cos\left(E_I(I_t), E_T(T_{c_t,a'})\right)\right)} \tag{25}$$

where $T_{c_t,a}$ is the context-conditioned prompt. However, absolute probability values from CLIP can be noisy and highly dependent on image complexity (Wang et al., 2024). Therefore, we focus on the relative confidence of the decision. We define a semantic neighbor set $\mathcal{N}(a_t)$ to exclude logically similar actions (e.g., "braking" vs. "braking hard") and calculate the reward margin. This margin quantifies the distinctiveness of the chosen action against the hardest valid negative specifically, the maximum probability among the remaining non-neighboring, competitive actions:

$$r_{\text{raw}} = \max\left(0, P(a_t \mid \cdot) - \max_{a_{\text{neg}} \notin \mathcal{N}(a_t)} P(a_{\text{neg}} \mid \cdot)\right) \tag{26}$$

Finally, we employ a Running Mean-Std (RMS) filter to normalize this margin. Since the VLM is pre-trained on vast expert datasets, a high alignment score serves as a proxy for "expertness" or "familiarity." By standardizing the raw margin against its running mean and standard deviation, the resulting signal acts as a fuzzy reward bonus that measures how much better the current action aligns with current state compared to the agent's average performance. The final shaped reward is computed by scaling this normalized bonus by $\lambda$ and adding it to the environmental reward $r_{env}$:

$$r_{\text{vlm}} = \text{clip}\left(\frac{r_{\text{raw}} - \mu_{\text{rms}}}{\sigma_{\text{rms}}}, -1, 1\right) \quad (27)$$

$$r_{\text{final}} = r_{env} + \lambda \cdot r_{\text{vlm}}$$

## Results

### 5.1. Experiment Setting

**Observation space and action space**

For the observation space, the VLM-based driving agents (Fig. 3 (a)) take a BEV image (192×192×3) as input, along with a text prompt. In contrast, the RL-based driving agents (Fig. 3 (b)) use BEV masks (96×96×15) together with a compact state vector (e.g., vehicle motion, last-step control signals, and traffic-rule context such as traffic lights and stop signs).

The action space is continuous. The VLM outputs 3D action (throttle, steer, brake), whereas the RL-based agents output 2D action (throttle/brake, steer), where throttle and brake are combined into a single signed longitudinal value (positive for throttle and negative for braking). In the VLM-enhanced RL method, the 3D VLM action is mapped into the 2D RL action accordingly.

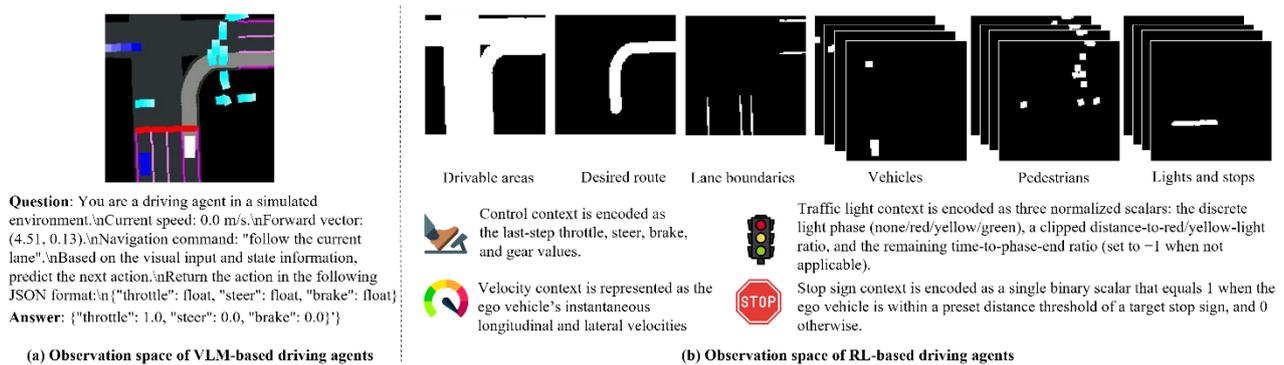

**Fig. 3.** Observation space.

**Reward functions and terminal criteria**

In this work, we adopt the ROACH-style reward shaping scheme (Zhang et al., 2021), where the total reward is defined as the sum of a speed-tracking term toward a desired speed (bounded by a fixed maximum speed), a route-keeping penalty based on lateral deviation from the reference centerline, a heading-alignment penalty based on angular error relative to the route direction, a smoothness penalty that discourages abrupt steering changes between consecutive steps, and a terminal reward/penalty. Following the same obstacle-aware rule (Toromanoff et al., 2019), the desired speed equals the maximum speed when no obstacle is detected and is linearly reduced to zero as the distance to a detected obstacle decreases.

For terminal criteria, an episode is terminated if the ego vehicle becomes blocked, violates the route constraint by exceeding an adaptive lateral-deviation threshold from the reference path, runs a red light, runs a stop sign, or triggers a collision event. In evaluation mode, termination further occurs when the route is completed. Compared to ROACH's evaluation setting, the protocol for evaluation applies a stricter and more training-consistent termination logic. In ROACH, evaluation termination is triggered by a small set of coarse events, which makes episodes less likely to end early. In contrast, our evaluation mode mirrors the training mode much more closely, including tighter route adherence checks and additional failure triggers, so episodes terminate more readily; consequently, while route completion can still be reasonable, the overall success rate tends to remain lower.

**Benchmarks**

For the Leaderboard benchmark, training is conducted on Town01, Town03, Town04, and Town06, and evaluation is performed on Town01, Town02, Town03, Town04, Town05, and Town06 (CARLA Team, n.d.). For the NoCrash benchmark, training is conducted on Town01, and evaluation is performed on Town01 and Town02 (Codevilla et al., 2019a). The map layouts are shown in Fig. 4.

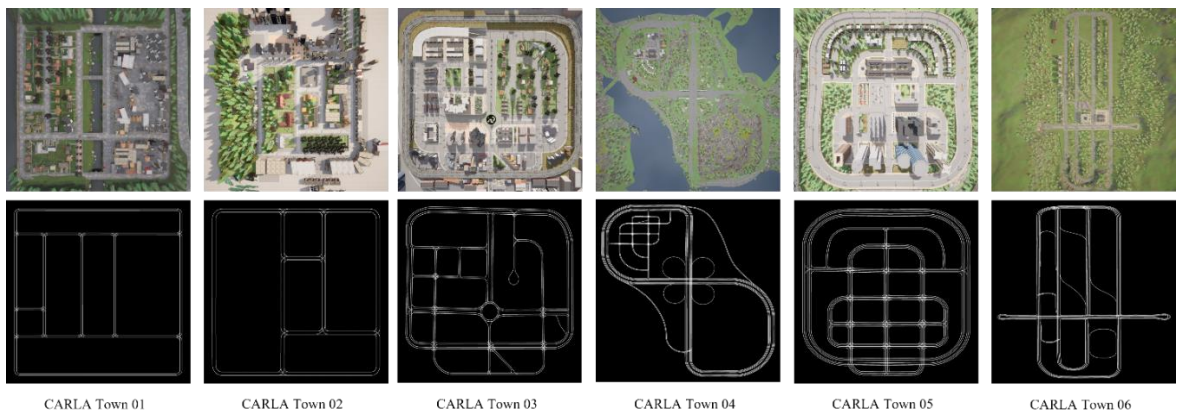

**Fig. 4.** Map layouts.

We additionally visualize representative and diverse scenarios from both benchmarks in Fig. 5, where white denotes the ego vehicle, dark blue indicates background vehicles, cyan indicates pedestrians, red/green denote traffic lights, dark gray denotes trajectories, light gray denotes the drivable area, and purple denotes road markings.

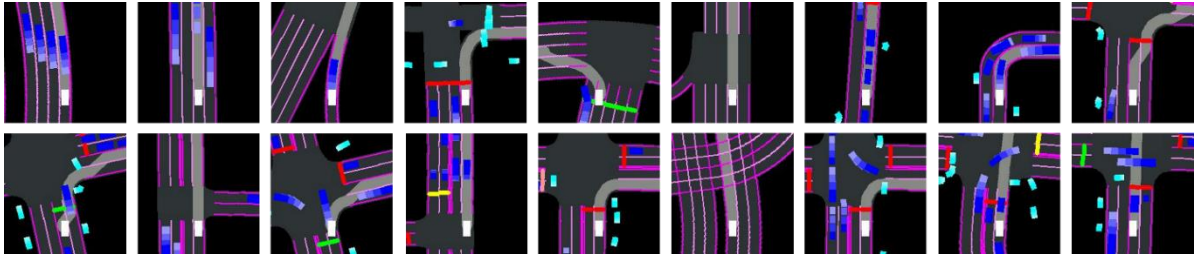

**Fig. 5.** Driving scenarios.

**Simulation Optimization**

To accommodate the partial observability of BEV inputs and enhance training stability, we implemented two key environmental adjustments. First, we observed a recurring artifact in BEV/BEV-mask settings: many static objects (e.g., poles, fences, and vegetation) are not represented in the observation, yet minor side contacts with these objects frequently trigger collision terminations. From the agent's perspective, such episodes often end with an "end state" that is nearly indistinguishable from the immediately preceding states, which introduces noisy credit assignment and substantially hampers early-stage learning. To remove this unobservable failure mode, we disable the physics of BEV-invisible static objects across all experiments. Importantly, this modification does not remove their functional roles when they are part of traffic logic—traffic lights and stop signs remain fully operational (e.g., phase control and rule enforcement) so that dynamic realism and compliance constraints are preserved. In addition, our reward design encourages centered lane keeping, making agents prefer driving near the lane centerline and further reducing the likelihood of boundary scraping behaviors. Second, to mitigate simulator instability caused by frequent re-initialization, we adopted a hybrid reset strategy. During early exploration (e.g., episode return < 100), we apply soft resets (teleporting the ego-vehicle) with a certain probability (e.g.,10 % or 25%) to reduce resource overhead, transitioning to hard resets later to ensure background actor vitality.

**Data collection**

We collected demonstration data on three benchmarks, including the CARLA Leaderboard benchmark (CARLA Team, n.d.), the NoCrash benchmark (Codevilla et al., 2019b), and the CARLA Challenge benchmark (CARLA Team, n.d.) to fine-tune VLMs. For the Leaderboard benchmark, we collected data from 160 episodes using the Roach PPO expert policy (Zhang et al., 2021), resulting in about 457k state-action transitions. For the NoCrash benchmark, we collected data from 80 episodes using the autopilot roaming expert policy, resulting in about 235k state-action transitions. For the CARLA Challenge benchmark, we collected data from 240 episodes using the autopilot roaming expert policy, resulting in about 682k state-action transitions. Across all benchmarks, this yields about 1.374 million transitions in total. For all benchmarks, the maximum duration of each episode was set to 300 seconds. The data logging further applies a terminal step filtering rule, where a short segment of steps immediately before a collision is not saved, so the dataset contains only the valid portion of each episode prior to the collision event. We use these data and open-source frameworks (Ilharco et al., 2021; Zheng et al., 2024) to fine-tune our VLMs.

**Metrics**

To comprehensively evaluate the performance of the driving agents, we use a set of metrics organized into four primary categories: **Comprehensive**, **Route**, **Energy**, and **Safety**. The Comprehensive metrics include *Return*, which represents the accumulated rewards over the entire episode; *Driving Score*, defined as the product of route completion and the infraction penalty; and *Infraction Penalty*, a discount factor aggregating all triggered infractions. Under Route performance, we report the *Success Rate* (the fraction of episodes successfully completed), *Route Completion* (the percentage of the route distance finished), and the average driving *Speed*. For Energy efficiency, we utilize *Icell* to estimate the battery cell current based on a single-cell equivalent circuit model, and *Fuel Rate* to measure instantaneous fuel consumption derived from a longitudinal road-load model. Finally, Safety is assessed by tracking the total number of *Collisions of Pedestrian*, *Collisions of Vehicle*, and *Red Light* violations over the completed route length.

*5.2. Baselines*

**Vision-Language Models.** We fine-tune three state-of-the-art VLM architectures on the full dataset collected from the three benchmarks (totaling 1.374M state-action transitions) to serve as both expert-action guides and standalone baselines. InternVL3 (Zhu et al., 2025) employs a native multimodal pre-training strategy with Variable Visual Position Encoding; we utilize the 1B and 2B variants. Qwen2.5-VL (Bai et al., 2025) supports dynamic resolution and absolute time encoding for precise temporal localization; we fine-tune the 3B and 7B models. Visual RWKV (Hou et al., 2025) is built on a linear recurrent backbone (Peng et al., 2023) that enables constant memory usage and efficient token-wise inference; we use the v7 0.1B variant.

**Online RLs.** We train standard online RL baselines using stable-baselines3 (Raffin et al., 2021). DrQ-v2 (Yarats et al., 2021) is a model-free algorithm optimized for pixel-based control via data-regularized Q-learning and image augmentation. SAC (Haarnoja et al., 2018) maximizes an entropy-augmented objective to encourage robust exploration, though high initial entropy can hinder efficiency in safety-critical tasks. TD3 (Fujimoto et al., 2018) mitigates overestimation bias and instability through clipped double Q-learning and delayed policy updates.

**Other methods.** We implement additional baselines including AD-domain methods, Imitation Learning (IL) and offline RLs. IL and offline RLs are trained using d3rlpy (Seno and Imai, 2022).

## 5.3. Compare with RLs & ablation analysis

To evaluate the effectiveness of the proposed Found-RL platform and the accompanying VLM supervision mechanisms, we conducted comprehensive experiments using the CARLA simulator. We benchmarked our approach against standard RL algorithms, including TD3, SAC, and DrQv2, to establish a performance baseline. All models were trained over three independent runs with different random seeds to ensure statistical reliability. The performance metrics are categorized into four distinct groups: Comprehensive, Route, Safety, and Energy, providing a holistic assessment of the driving agents.

Table 1 presents the comparative results on the Leaderboard benchmark, demonstrating that integrating VLM feedback significantly enhances performance across all metric categories. In terms of Comprehensive and Route capabilities, the proposed methods consistently outperform baselines; notably, DrQv2-CLIP achieves state-of-the-art results with a Driving Score of 0.77 and a Success Rate of 57%, substantially surpassing the vanilla DrQv2 baseline. Crucially, this aggressive progress does not compromise Safety, as VLM-enhanced agents exhibit reduced collision rates—for example, SAC-VMR lowers vehicle collisions from 0.50 to 0.20 per km compared to SAC. Furthermore, the methods maintain high Energy efficiency, with SAC-VMR achieving the lowest battery consumption (Icell =0.07), proving that VLM-derived supervision effectively guides agents toward policies that are not only robust and safe but also smooth and energy-efficient.

**Table 1.** Compare with RLs on Leaderboard benchmark

| Algorithm | Comprehensive | | | | | Route | Energy | | Safety | | |
|---|---|---|---|---|---|---|---|---|---|---|---|
| | Return ↑ | Driving Score ↑ | Infra. Penalty ↑ | Success Rate ↑ | Route Compl. ↑ | Speed ↑ | Icell ↓ | Fuel Rate ↓ | Collisions Ped. ↓ | Collisions Veh. ↓ | Red Light ↓ |
| TD3 | 1703.83±843.48 | 0.53±0.24 | 0.89±0.03 | 0.37±0.30 | 0.58±0.26 | 3.48±0.17 | 0.18±0.01 | 0.004±0.000 | 0.00±0.00 | 0.21±0.07 | 0.03±0.02 |
| SAC | 527.03±102.66 | 0.21±0.06 | 0.90±0.04 | 0.03±0.02 | 0.23±0.06 | 3.73±0.09 | 0.15±0.02 | 0.003±0.000 | 0.01±0.02 | 0.50±0.35 | 0.05±0.03 |
| SAC-AWAG | 556.33±152.82 | 0.28±0.05 | 0.82±0.01 | 0.08±0.04 | 0.33±0.06 | **5.21±0.10** | 0.08±0.01 | 0.002±0.000 | 0.01±0.01 | 0.63±0.09 | 0.15±0.09 |
| SAC-VMR | 1623.34±265.96 | 0.53±0.10 | 0.89±0.03 | 0.31±0.14 | 0.58±0.11 | 3.63±0.27 | 0.07±0.01 | 0.002±0.000 | 0.00±0.00 | 0.20±0.05 | 0.02±0.02 |
| DrQv2 | 1507.66±517.97 | 0.56±0.11 | 0.88±0.01 | 0.38±0.13 | 0.61±0.12 | 3.72±0.20 | 0.09±0.03 | 0.002±0.001 | 0.01±0.01 | 0.19±0.01 | 0.02±0.03 |
| DrQv2-AWAG | 1626.35±497.37 | 0.61±0.10 | **0.90±0.02** | 0.46±0.15 | 0.66±0.10 | 3.68±0.27 | **0.06±0.01** | **0.001±0.000** | 0.01±0.01 | **0.13±0.03** | 0.02±0.02 |
| DrQv2-VMR | **2237.31±173.09** | 0.72±0.04 | 0.89±0.03 | **0.60±0.04** | **0.78±0.05** | 3.56±0.09 | 0.09±0.05 | 0.002±0.001 | 0.00±0.00 | 0.16±0.04 | 0.02±0.01 |
| DrQv2-CLIP | 2188.54±204.03 | **0.77±0.05** | **0.90±0.02** | 0.57±0.06 | 0.77±0.05 | 3.60±0.10 | 0.13±0.03 | 0.002±0.002 | **0.00±0.00** | 0.14±0.05 | **0.01±0.00** |

Table 2 and Table 3 detail performance on the NoCrash benchmark across Town01 (the training town of NoCrash benchmark) and Town02 (an unseen town of NoCrash benchmark), collectively demonstrating that VLM-enhanced RL not only stabilizes weaker baselines in familiar environments but also fosters significant generalization capabilities in novel scenarios. In terms of comprehensive driving metrics, the impact is transformative for the SAC family; SAC-VMR elevates the Driving Score in Town01 from a negligible 0.24 to a competitive 0.71, and substantially recovers performance in the unseen Town02, boosting the Success Rate from a failing 7% to 47%. Meanwhile, the strong DrQv2-CLIP baseline maintains peak performance in the training environment and achieves the highest overall Driving Score of 0.76 in Town02, confirming its resilience to increased traffic complexity. Crucially, VLM feedback proves vital for collision avoidance and efficiency across both domains. In Town01, SAC-VMR significantly reduces the high vehicle collision rate of the vanilla agent from 1.81 to 0.59 per km, while in the more challenging Town02, DrQv2-CLIP nearly halves the baseline's rate (reducing it from 1.27 to 0.73 per km), confirming that the contrastive alignment reward successfully encodes transferable obstacle-awareness. Finally, the energy analysis highlights consistent efficiency gains from action guidance, with SAC-AWAG achieving remarkable energy conservation ($I_{cell} \approx 0.04$) compared to baselines ($> 0.11$), demonstrating that VLM-guided policies promote smoother, less aggressive actuation alongside their safety benefits.

**Table 2.** Compare with RLs on NoCrash benchmark (Town01)

| Algorithm | Comprehensive | | | | Route | | Energy | | Safety | | |
|---|---|---|---|---|---|---|---|---|---|---|---|
| | Return ↑ | Driving Score ↑ | Infra. Penalty ↑ | Success Rate ↑ | Route Compl. ↑ | Speed ↑ | Icell ↓ | Fuel Rate ↓ | Collisions Ped. ↓ | Collisions Veh. ↓ | Red Light ↓ |
| TD3 | 817.82±113.51 | 0.45±0.04 | 0.80±0.04 | 0.20±0.14 | 0.52±0.08 | 2.78±0.22 | 0.13±0.02 | 0.003±0.000 | 0.01±0.02 | 0.68±0.15 | 0.07±0.01 |
| SAC | 640.28±312.40 | 0.24±0.11 | 0.87±0.01 | 0.04±0.06 | 0.27±0.12 | 2.50±0.51 | 0.11±0.06 | 0.003±0.001 | 0.00±0.00 | 1.81±0.44 | 0.07±0.06 |
| SAC-AWAG | 995.53±186.42 | 0.50±0.04 | 0.80±0.04 | 0.30±0.07 | 0.60±0.04 | **3.18±0.14** | **0.04±0.00** | **0.001±0.000** | 0.00±0.00 | 1.10±0.26 | 0.13±0.08 |
| SAC-VMR | 1639.04±33.76 | 0.71±0.05 | 0.85±0.03 | 0.56±0.06 | 0.81±0.02 | 2.70±0.08 | 0.05±0.01 | 0.001±0.000 | 0.00±0.00 | 0.59±0.07 | 0.09±0.03 |
| DrQv2 | **1658.44±126.95** | 0.73±0.06 | 0.86±0.02 | **0.66±0.12** | **0.84±0.08** | 2.75±0.14 | 0.10±0.06 | 0.002±0.001 | 0.00±0.00 | 0.56±0.06 | 0.05±0.07 |
| DrQv2-AWAG | 1393.96±345.09 | 0.74±0.10 | **0.91±0.02** | 0.54±0.21 | 0.79±0.11 | 2.57±0.34 | 0.06±0.04 | 0.001±0.001 | 0.01±0.01 | 0.34±0.10 | **0.04±0.02** |
| DrQv2-VMR | 1448.10±144.54 | 0.70±0.03 | 0.88±0.03 | 0.56±0.09 | 0.80±0.05 | 2.92±0.13 | 0.11±0.06 | 0.002±0.001 | 0.00±0.00 | 0.48±0.14 | 0.04±0.01 |
| DrQv2-CLIP | 1535.69±324.61 | **0.75±0.05** | 0.90±0.03 | 0.65±0.05 | 0.81±0.04 | 2.51±0.09 | 0.09±0.04 | 0.002±0.000 | **0.00±0.00** | 0.32±0.17 | 0.07±0.07 |

**Table 3.** Compare with RLs on NoCrash benchmark (Town02)

| Algorithm | Comprehensive | | | | Route | | Energy | | Safety | | |
|---|---|---|---|---|---|---|---|---|---|---|---|
| | Return ↑ | Driving Score ↑ | Infra. Penalty ↑ | Success Rate ↑ | Route Compl. ↑ | Speed ↑ | Icell ↓ | Fuel Rate ↓ | Collisions Ped. ↓ | Collisions Veh. ↓ | Red Light ↓ |
| TD3 | 463.76±116.39 | 0.39±0.10 | 0.78±0.06 | 0.20±0.11 | 0.47±0.14 | 2.08±0.27 | 0.11±0.01 | 0.002±0.000 | 0.00±0.00 | 1.94±0.11 | 0.13±0.07 |
| SAC | 576.45±381.10 | 0.24±0.16 | 0.91±0.02 | 0.07±0.12 | 0.25±0.17 | 1.76±0.48 | 0.08±0.04 | 0.002±0.001 | 0.00±0.00 | 2.44±0.98 | 0.44±0.32 |
| SAC-AWAG | 474.22±178.52 | 0.46±0.04 | 0.79±0.03 | 0.29±0.07 | 0.56±0.04 | **2.83±0.32** | **0.05±0.01** | **0.001±0.000** | 0.00±0.00 | 2.38±0.72 | 0.63±0.35 |
| SAC-VMR | 819.41±105.87 | 0.61±0.03 | 0.84±0.04 | 0.47±0.04 | 0.70±0.05 | 2.24±0.10 | 0.06±0.01 | 0.001±0.000 | 0.00±0.00 | 1.32±0.21 | 0.27±0.15 |
| DrQv2 | 1023.42±121.77 | 0.73±0.03 | 0.83±0.03 | **0.77±0.06** | **0.86±0.00** | 2.07±0.12 | 0.08±0.03 | 0.002±0.001 | 0.01±0.02 | 1.27±0.33 | **0.04±0.04** |
| DrQv2-AWAG | 830.53±256.79 | 0.69±0.11 | 0.87±0.07 | 0.63±0.21 | 0.79±0.15 | 2.02±0.38 | 0.06±0.03 | 0.001±0.001 | 0.00±0.00 | 0.90±0.64 | 0.10±0.13 |
| DrQv2-VMR | **1060.89±115.57** | 0.64±0.02 | 0.79±0.05 | 0.69±0.11 | 0.81±0.07 | 2.21±0.11 | 0.08±0.03 | 0.002±0.001 | 0.03±0.03 | 1.51±0.37 | 0.07±0.03 |
| DrQv2-CLIP | 953.98±116.80 | **0.76±0.09** | 0.87±0.06 | 0.71±0.11 | 0.85±0.04 | 1.85±0.28 | 0.08±0.02 | 0.002±0.000 | **0.00±0.00** | **0.73±0.51** | 0.07±0.04 |

We further analyze the training dynamics and sample efficiency in Fig. 6. As the agents are trained in an endless environment, we report the Average Reward and Route Completed on a per-step basis rather than per episode. The curves illustrate that incorporating VLM feedback significantly accelerates the learning process, particularly in the early training stages. For instance, in the SAC-based experiments (columns c and d), the VLM-guided variants (AWAG and VMR) exhibit a steep initial ascent in both reward acquisition and route progression, rapidly distancing themselves from the slower-learning baseline. While the strong DrQv2 baseline remains competitive, the VLM-enhanced methods, especially DrQv2-CLIP, demonstrate superior asymptotic performance and reduced variance. Ultimately, the plateauing trends across all subplots confirm that the proposed algorithms successfully converge to stable, high-performing policies, validating that the integration of semantic supervision expedites the discovery of optimal control strategies without destabilizing the learning process.

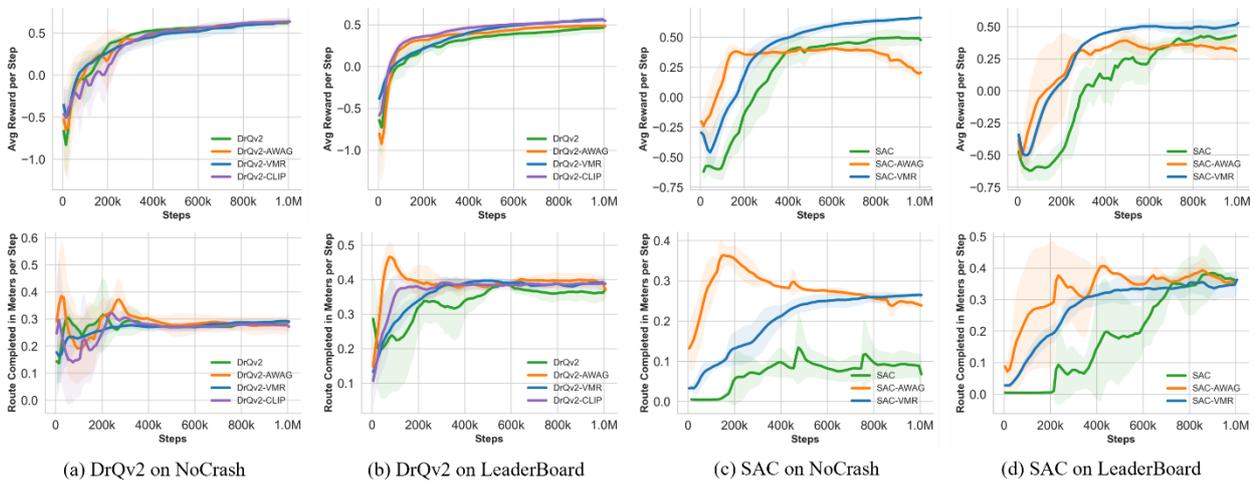

**Fig. 6.** Comparison of average reward and route completion in meters per step between Found-RL and classical RLs on Leaderboard and NoCrash benchmarks.

Fig. 7 illustrates the evolution of actor and critic losses throughout the training process, offering insight into the internal optimization dynamics of the proposed methods. A distinct characteristic is observed in the AWAG variants (orange curves), where the Actor loss remains exceptionally flat and stable compared to the baselines; this stability is intrinsic to the AWAG formulation, where the VLM guidance term is adaptively scaled by the inverse of the Q-value magnitude $\left(\frac{1}{|Q|}\right)$ to normalize gradient magnitudes and prevent the auxiliary supervision from overwhelming the primary RL objective. In contrast, the VMR method (blue curves) exhibits a characteristic "dip-and-recover" dynamic driven by its margin-based objective. Mathematically, the auxiliary VMR loss forces the critic to assign a higher value to VLM-proposed actions than to the policy's actions by a fixed margin $\Delta$, systematically inflating Qvalues in the early stages. Since the actor's objective is to maximize these values (minimizing $-Q$), this inflation causes the actor loss to drop sharply, creating the deep valleys observed in the curves. As the regularization coefficient $\lambda$ undergoes cosine decay and the margin constraint relaxes, the Q-values gradually realign with true environmental returns, allowing the loss trajectory to recover and stabilize. Ultimately, both actor and critic losses for all VLM-enhanced algorithms converge to a steady state, confirming that the asynchronous integration of semantic feedback provides a robust learning signal without inducing divergence or long-term instability.

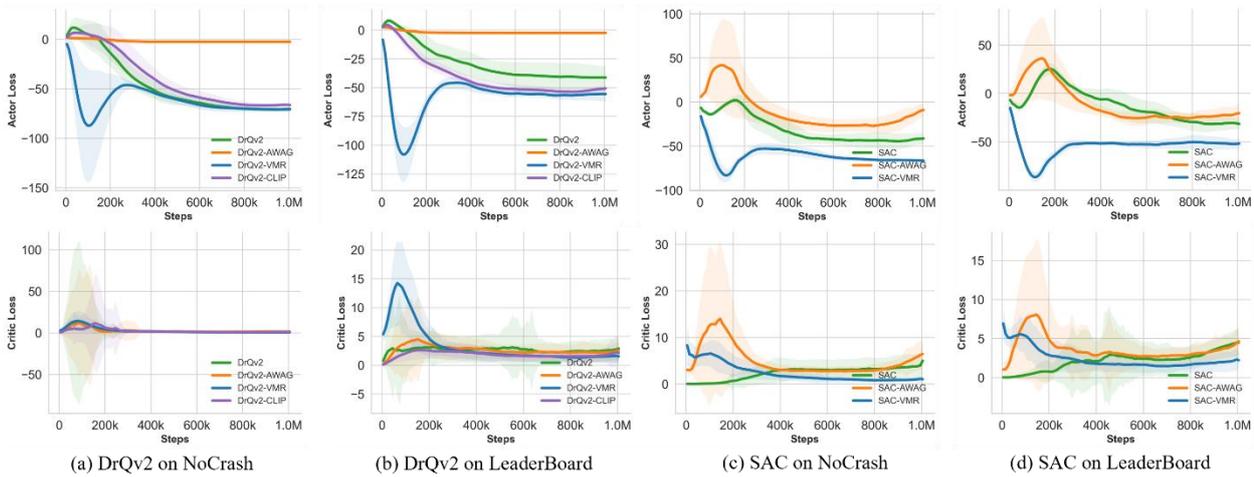

**Fig. 7.** Comparison of actor and critic losses between Found-RL and classical RLs on Leaderboard and NoCrash benchmarks.

Fig. 8 depicts the evolution of the auxiliary VMR and AWAG losses, effectively illustrating the dynamic trade-off between VLM-guided imitation and autonomous self-exploration. For most configurations, such as DrQv2-AWAG and DrQv2-VMR, we employ a decay schedule for the guidance coefficient (marked by the vertical dashed line). Initially, the losses are low as the agent closely mimics the VLM's priors. However, as the coefficient decays and the "Self-exploration" phase begins, the auxiliary loss naturally rises; this trend is desirable, indicating that the agent is no longer strictly bound by the teacher's behavior and is successfully exploring the environment to maximize the primary RL objective. A notable exception is observed in the SAC-AWAG variant (Fig. 8 (c)), where the loss remains stable and near-zero throughout training. In this specific setting, we maintained a constant guidance coefficient without decay, as empirical observations revealed that relaxing the VLM constraint in the early stages led to performance instability and policy collapse. This constant regularization ensures that the SAC agent remains firmly anchored to the semantic safety constraints provided by the VLM throughout the entire learning process.

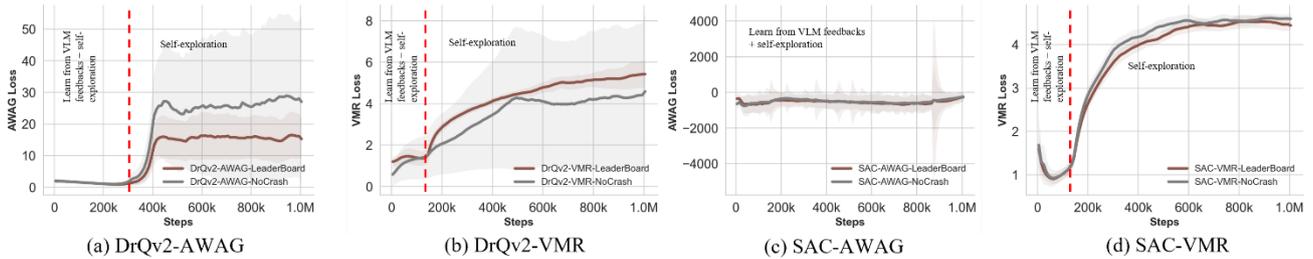

**Fig. 8.** Comparison of AWAG and VMR losses between Found-RL and classical RLs on Leaderboard and NoCrash benchmarks.

### *5.4. Compare with VLMs*

Table 4 benchmarks our foundation model-enhanced RL agents against VLMs on the Leaderboard benchmark, demonstrating that our agents achieve highly competitive performance despite vast differences in scale. In terms of comprehensive metrics, DrQv2-CLIP attains a Driving Score of 0.77, which not only significantly outperforms the 1B-parameter InternVL3 (0.63) and RWKV (0.65) baselines but also edges out the 7B-parameter Qwen2.5-vl (0.76). Furthermore, the RL agents exhibit superior fine-grained control in safety metrics, maintaining exceptionally low red light violation rates (0.01–0.02 per km) compared to the VLM baselines (0.04–0.08). Notably, these results are achieved using a lightweight model of only 3.82 M parameters and reduced $96\times96$ visual inputs (only 25% of the spatial size used by VLMs), effectively rivaling the performance of multi-billion parameter models that process significantly larger $192\times192$ inputs.

**Table 4.** Compare with VLMs on Leaderboard benchmark

| Algorithm | Model Size | Input Modality | Comprehensive | | | Route | | | Energy | | Safety | | |
|---|---|---|---|---|---|---|---|---|---|---|---|---|---|
| | | | Return ↑ | Driving Score ↑ | Infra. Penalty ↑ | Succ. Rate ↑ | Route Compl. ↑ | Speed ↑ | Icell ↓ | Fuel Rate ↓ | Coll. Ped. ↓ | Coll. Veh. ↓ | Red Light ↓ |
| Visual | 0.1B | BEV (192×192×3) | 2044.01 | 0.65 | 0.86 | 0.52 | 0.70 | 3.63 | 0.05 | 0.001 | 0.01 | 0.15 | 0.08 |

| Algorithm | Model Size | Input Modality | Comprehensive Return ↑ | Driving Score ↑ | Infra. Penalty ↑ | Succ. Rate ↑ | Route Compl. ↑ | Speed ↑ | Energy Icell ↓ | Fuel Rate ↓ | Coll. Ped. ↓ | Coll. Veh. ↓ | Red Light ↓ |
|---|---|---|---|---|---|---|---|---|---|---|---|---|---|
| RWKV 0B1 | | | | | | | | | | | | | |
| Qwen2.5-vl-7b | 7B | BEV (192×192×3) | 2702.49 | 0.76 | 0.91 | 0.65 | 0.80 | 3.50 | 0.04 | 0.001 | 0.00 | 0.09 | 0.05 |
| Qwen2.5-vl-3b | 3B | BEV (192×192×3) | 2763.70 | 0.74 | 0.90 | 0.59 | 0.79 | 3.38 | **0.03** | 0.001 | 0.00 | 0.13 | 0.05 |
| Internvl3-2b | 2B | BEV (192×192×3) | **3185.72** | **0.85** | **0.92** | **0.75** | **0.89** | 3.40 | **0.03** | 0.001 | 0.00 | **0.08** | 0.04 |
| Internvl3-1b | 1B | BEV (192×192×3) | 2157.70 | 0.63 | 0.87 | 0.49 | 0.68 | 3.83 | 0.04 | 0.001 | 0.00 | 0.19 | 0.07 |
| DrQv2-VMR | 3.82 M | BEV Masks (96×96×15) + State (10, ) | 2237.31 | 0.72 | 0.89 | 0.62 | 0.78 | 3.56 | 0.09 | 0.002 | 0.00 | 0.16 | 0.02 |
| DrQv2-CLIP | 3.82 M | BEV Masks (96×96×15) + State (10, ) | 2188.54 | 0.77 | 0.90 | 0.57 | 0.77 | 3.60 | 0.13 | 0.002 | **0.00** | 0.14 | **0.01** |

Table 5 and Table 6 extend the comparative analysis to the NoCrash benchmarks, revealing a nuanced trade-off between model capacity and efficiency. While the large-scale VLMs (1B–7B parameters) generally achieve higher upper-bound Driving Scores, peaking at 0.92 in Town01 and 0.94 in Town02, our lightweight RL agents remain highly competitive against comparable baselines. Specifically, DrQv2-CLIP achieves Success Rates of 0.65 (Town01) and 0.71 (Town02), consistently outperforming the Visual RWKV baseline (0.61 and 0.63, respectively). Objectively, the RL agents exhibit higher vehicle collision rates than the multi-billion parameter models, likely a limitation of the reduced 96×96 visual resolution in handling complex dynamic agents; however, they maintain superior adherence to static traffic rules, recording significantly lower red light violation rates (e.g., 0.07 per km in Town02) compared to even powerful VLMs like Qwen2.5-vl-7b (0.29).

**Table 5.** Compare with VLMs on NoCrash benchmark (Town01)

| Algorithm | Model Size | Input Modality | Comprehensive Return ↑ | Driving Score ↑ | Infra. Penalty ↑ | Succ. Rate ↑ | Route Compl. ↑ | Speed ↑ | Energy Icell ↓ | Fuel Rate ↓ | Coll. Ped. ↓ | Coll. Veh. ↓ | Red Light ↓ |
|---|---|---|---|---|---|---|---|---|---|---|---|---|---|
| Visual RWKV 0B1 | 0.1B | BEV (192×192×3) | 1588.76 | 0.82 | 0.91 | 0.61 | 0.87 | 2.61 | 0.03 | 0.001 | 0.00 | 0.32 | 0.11 |
| Qwen2.5-vl-7b | 7B | BEV (192×192×3) | 1731.70 | **0.92** | **0.97** | **0.86** | **0.94** | 2.73 | **0.03** | **0.001** | 0.02 | **0.04** | 0.08 |
| Qwen2.5-vl-3b | 3B | BEV (192×192×3) | 1709.61 | 0.85 | 0.96 | 0.75 | 0.87 | 2.56 | 0.03 | 0.001 | 0.00 | 0.08 | 0.13 |
| Internvl3-2b | 2B | BEV (192×192×3) | 1851.79 | 0.85 | 0.94 | 0.74 | 0.88 | 2.38 | 0.03 | 0.001 | 0.00 | 0.15 | 0.15 |
| Internvl3-1b | 1B | BEV (192×192×3) | **1922.43** | 0.89 | 0.96 | 0.68 | 0.91 | 2.58 | 0.03 | 0.001 | 0.00 | 0.10 | 0.08 |
| DrQv2-VMR | 3.82 M | BEV Masks (96×96 | 1448.10 | 0.70 | 0.88 | 0.56 | 0.80 | **2.92** | 0.11 | 0.002 | **0.00** | 0.48 | **0.04** |

| Algorithm | Model Size | Input Modality | Comprehensive Return ↑ | | | | | | | | | |
|---|---|---|---|---|---|---|---|---|---|---|---|---|
| DrQv2-CLIP | 3.82 M | BEV Masks (96×96 ×15) + State (10, ) | 1535.69 | 0.75 | 0.90 | 0.65 | 0.81 | 2.51 | 0.09 | 0.002 | 0.00 | 0.32 | 0.07 |

Table 6. Compare with VLMs on NoCrash benchmark (Town02)

| Algorithm | Model Size | Input Modality | Comprehensive | | Route | | | Energy | | Safety | | |
|---|---|---|---|---|---|---|---|---|---|---|---|---|
| | | | Return ↑ | Driving Score ↑ | Infra. Penalty ↑ | Succ. Rate ↑ | Route Compl. ↑ | Speed ↑ | Icell ↓ | Fuel Rate ↓ | Coll. Ped. ↓ | Coll. Veh. ↓ | Red Light ↓ |
| Visual RWKV 0B1 | 0.1B | BEV (192×192×3) | 768.25 | 0.86 | 0.94 | 0.63 | 0.89 | 1.54 | **0.02** | 0.001 | 0.00 | 0.30 | 0.30 |
| Qwen2.5-vl-7b | 7B | BEV (192×192×3) | 1132.63 | 0.89 | 0.96 | 0.83 | 0.91 | 2.03 | 0.03 | 0.001 | 0.00 | **0.07** | 0.29 |
| Qwen2.5-vl-3b | 3B | BEV (192×192×3) | 947.57 | **0.94** | **0.97** | **0.91** | **0.95** | 2.18 | 0.03 | **0.001** | 0.08 | 0.12 | **0.00** |
| Internvl3-2b | 2B | BEV (192×192×3) | **1166.86** | 0.91 | 0.96 | 0.85 | 0.93 | 2.02 | 0.03 | 0.001 | 0.00 | 0.17 | 0.13 |
| Internvl3-1b | 1B | BEV (192×192×3) | 1005.63 | 0.85 | 0.92 | 0.60 | 0.89 | 1.97 | 0.03 | 0.001 | 0.00 | 0.52 | 0.16 |
| DrQv2-VMR | 3.82 M | BEV Masks (96×96 ×15) + State (10, ) | 1060.89 | 0.64 | 0.79 | 0.69 | 0.81 | **2.21** | 0.08 | 0.002 | 0.03 | 1.51 | 0.07 |
| DrQv2-CLIP | 3.82 M | BEV Masks (96×96 ×15) + State (10, ) | 953.98 | 0.76 | 0.87 | 0.71 | 0.85 | 1.85 | 0.08 | 0.002 | **0.00** | 0.73 | 0.07 |

In conclusion, the comparative evaluation across both Leaderboard and NoCrash benchmarks demonstrates that our foundation model-enhanced RL agents achieve driving proficiency comparable to large-scale VLMs while operating under drastically tighter resource constraints. Despite the massive disparity in scale, our DrQv2-CLIP agent delivers competitive Success Rates and superior rule compliance (e.g., minimal red light violations) using a lightweight model of only 3.82 M parameters. This stands in stark contrast to the 1B–7B parameter VLM baselines, representing a model size reduction of approximately 260x to 1800x. Furthermore, our approach achieves these results using compressed $96\times 96$ visual inputs—a 75% reduction in spatial resolution compared to the $192\times 192$ inputs required by VLMs. These results validate that our framework effectively distills the semantic reasoning of foundation models into compact, efficient policies, offering a viable solution for deploying high-performance autonomous driving agents on resource-constrained platforms.

## 5.4. Compare with other baselines

To strictly evaluate performance (Table 7), we compare Found-RL against two categories of baselines: AD-domain methods that utilize high-fidelity RGB and LiDAR inputs, and general-purpose Imitation Learning (IL)/offline RL algorithms operating on lightweight BEV maps. It is important to note that this comparison is conducted under asymmetric conditions; while AD-domain methods' results are derived from official Leaderboard that allow continued driving after minor infractions, our evaluation enforces stricter safety-critical termination criteria to ensure trustworthy behavior. Although this rigorous protocol naturally constrains Route Completion by terminating episodes early upon unsafe deviations, Found-RL still achieves the highest Driving Score (0.77) with a superior Infraction Penalty (0.90). This demonstrates that despite relying on significantly lower-dimensional BEV inputs, our RL paradigm of learning from VLM feedback fosters robust and compliant driving policies that prioritize safety over the aggressive trajectory tracking often observed in standard baselines.

**Table 7.** Compare with other baselines on Leaderboard benchmark. † denotes results cited from the official Leaderboard. Unlike these baselines, our method uses stricter termination criteria and compact inputs, achieving the highest Driving Score through superior rule compliance rather than aggressive driving.

| Catrgories | Algorithms | Input Modality | Driving Score ↑ | Route Compl. ↑ | Infra. Penalty ↑ |
|---|---|---|---|---|---|
| AD-domain methods † | NEAT (Chitta et al., 2021) | RGB | 0.22 | 0.42 | 0.65 |
| | IARL (Toromanoff et al., 2020) | RGB | 0.25 | 0.47 | 0.52 |
| | Rails (Chen et al., 2021) | RGB | 0.31 | 0.58 | 0.56 |
| | Latent TransFuser (Chitta et al., 2022) | RGB + LiDAR | 0.45 | 0.66 | 0.72 |
| | GRIAD (Chekroun et al., 2023) | RGB + LiDAR | 0.37 | 0.62 | 0.60 |
| | TransFuser (Chitta et al., 2022) | RGB + LiDAR | 0.61 | 0.87 | 0.71 |
| | LAV (Chen and Krähenbühl, 2022) | RGB + LiDAR | 0.62 | **0.94** | 0.64 |
| | TCP (Wu et al., 2022) | RGB | 0.75 | 0.86 | 0.87 |
| | InterFuser (Shao et al., 2023) | RGB + LiDAR | 0.76 | 0.88 | 0.84 |
| IL and offline RLs | BC | BEV | 0.36 | 0.40 | 0.89 |
| | AWAC (Nair et al., 2020) | BEV | 0.29 | 0.37 | 0.75 |
| | TD3-BC (Fujimoto and Gu, 2021) | BEV | 0.29 | 0.37 | 0.74 |
| | CQL (Kumar et al., 2020) | BEV | 0.15 | 0.18 | 0.87 |
| | IQL (Kostrikov et al., 2021) | BEV | 0.20 | 0.27 | 0.75 |
| VLM-enhanced RL | Found-RL (DrQv2-CLIP, ours) | BEV Masks + State | **0.77** | 0.77 | **0.90** |

## 5.5. Analysis on CLIP Scoring

To quantitatively assess the discriminative capability of the fine-tuned CLIP model within the proposed semantic action space, we performed a retrieval-based classification experiment on a held-out evaluation subset comprising 20,000 samples. For each visual observation, we constructed 30 candidate text prompts representing the full spectrum of lateral and longitudinal behaviors, all conditioned on the corresponding ground-truth navigation command and speed state to isolate the action recognition performance. The predicted action was determined by identifying the specific text anchor among the candidates that maximized the cosine similarity to the image embedding, thereby testing the model's ability to disambiguate fine-grained driving maneuvers solely from visual cues.

The resulting row-normalized confusion matrix as shown in Fig. 9 exhibits a pronounced diagonal structure, confirming that the model effectively disentangles distinct driving maneuvers despite the high degree of visual similarity between adjacent frames in the bird's-eye view. Crucially, the off-diagonal predictions are not uniformly distributed but are heavily concentrated within local semantic clusters. Misclassifications predominantly occur between intensity variations of the same maneuver, such as identifying "braking hard" as "braking" or "accelerating fast" as "accelerating", reflecting the inherent ambiguity in discretizing continuous dynamics rather than fundamental perception failures. Conversely, the model exhibits clear separation between conflicting directional maneuvers, with negligible confusion observed between opposing lateral actions like turning left versus turning right. This pattern indicates that the learned latent space preserves the semantic topology of driving behaviors, where prediction deviations correspond to logical near-misses. This characteristic explicitly validates our reward formulation strategy, confirming that the semantic neighbor masking is necessary to accommodate these reasonable linguistic ambiguities while the underlying representation remains robust and directionally accurate.

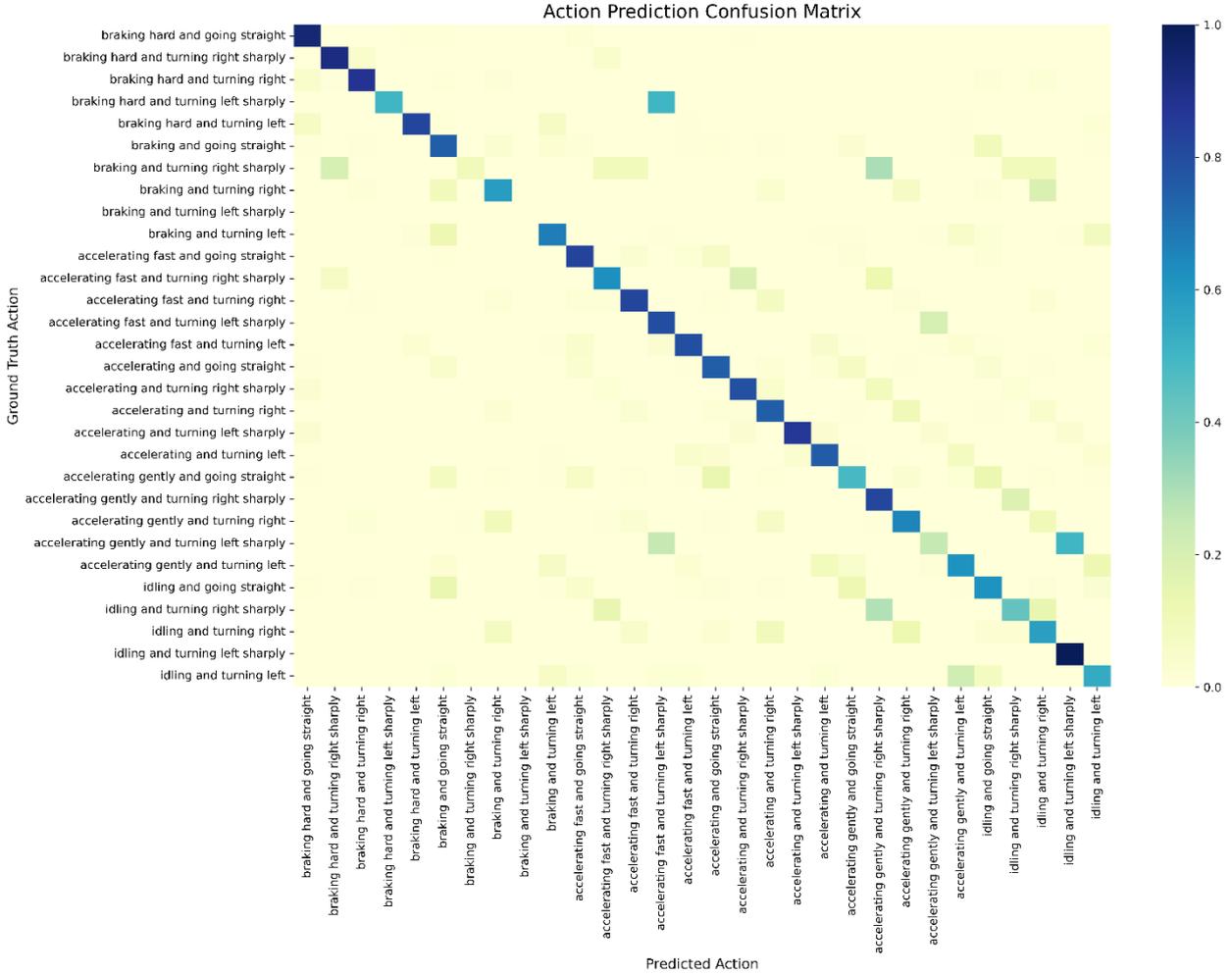

**Fig. 9.** Action prediction confusion matrix.

We analyze the system feasibility and training dynamics in Fig. 10. The left panel demonstrates that the reward availability and consistently remains near 100%, a critical metric since dense reward shaping necessitates a guiding signal for every transition; this high throughput explicitly justifies our choice of the efficient CLIP encoder over computationally heavier generative VLMs that would bottleneck the simulation. The right panel depicts the evolution of the average semantic alignment margin (Eq 26) across the entire replay buffer, where the initial upward trend indicates the agent effectively learning to align its visual behavior with safety prompts, while the plateau after approximately 300k steps suggests the agent has converged on coarse-grained safety behaviors by eliminating gross semantic errors, such as braking during acceleration commands or turning opposite to the navigation goal, thereby maintaining a consistently high safety margin.

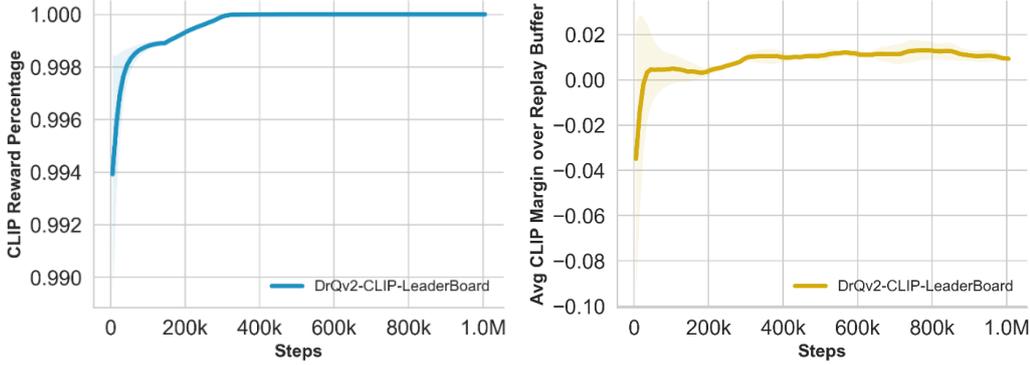

**Fig. 10.** Reward percentage and average CLIP margin over replay buffer.

*5.6. Efficiency analysis*

FPS is evaluated in an offline setting, calculated as the average rate over 1,000 consecutive inference runs. This metric highlights the substantial performance advantage shown in Table 8 where Found-RL, as a lightweight reinforcement learning-based solution, demonstrates superior real-time capabilities compared to Vision-Language Models (VLMs). With a compact size of only 3.82M parameters, Found-RL achieves a remarkable 500 FPS, standing in stark contrast to the 1B to 7B parameter VLMs that operate at approximately 1 FPS. This speed difference of several orders of magnitude confirms that Found-RL's small-scale architecture is uniquely suited for high-frequency, real-time inference tasks that are computationally prohibitive for VLMs.

Additionally, VLM inference speed does not strictly scale with model size. Qwen2.5-vl-3b is notably slower than its 7B counterpart, likely due to its specific layer number configuration. Furthermore, Visual RWKV 0.1B exhibits the lowest FPS despite being the smallest model, a discrepancy attributed to its unique architecture or the high computational overhead of its visual encoder.

**Table 8.** Comparison between Found-RL and VLMs on FPS and model size

|  | Found-RL (ours) | Internvl3-1b | Internvl3-2b | Qwen2.5-vl-3b | Qwen2.5-vl-7b | Visual RWKV-0.1b |
|---|---|---|---|---|---|---|
| FPS | **500** | 1.26 | 1.09 | 0.74 | 0.89 | 0.44 |
| Model Size | **3.82M** | 1B | 2B | 3B | 7B | 0.1B |

## Conclusions

In this paper, we proposed a novel framework that effectively integrates Foundation Models/VLMs into the RL loop, providing both dense reward shaping and explicit action guidance to address sample inefficiency and safety challenges in autonomous driving. We demonstrated that our method not only accelerates learning compared to standard RL baselines but also achieves performance comparable to computationally intensive VLMs, despite utilizing significantly smaller spatial inputs, less training data, and a compact model size (3.82M parameters) in contrast to massive 0.1B-7B parameter models. While our asynchronous batch inference framework enables efficient background reasoning, ensuring that VLM feedback remains synchronized with rapid RL training is crucial; thus, future work will explore inference acceleration and quantization to prevent stale feedback and ensure broad sample coverage when scaling to larger models. Additionally, we aim to validate our robustness on more challenging benchmarks like CARLA Leaderboard 2.0, and envision orchestrating a multi-VLM ensemble where distinct models provide specialized feedback on different driving aspects to collaboratively enhance learning efficiency, or even integrating lightweight VLMs directly as policy backbones for end-to-end planning.

## Appendix

### Appendix A. Fine-tuning CLIP and discretion scheme

We discretize vehicle speed and control actions according to the semantic thresholds defined in Table 9, Table 10, and Table 11, while navigation goals are represented by six standard high-level commands (e.g., turn left at the intersection, turn right at the intersection, go straight at the intersection, follow the current lane, change to the left lane, change to the right lane). Using this discretized schema and the collected data, we fine-tune an OpenCLIP (Ilharco et al., 2021) ViT-B-16 model (initialized from laion2b_s34b_b88k) for 10 epochs to align these visual observations with the synthesized state-action captions.

**Table 9.** Speed discretion

| Speed | Text Prompt |
|---|---|
| speed < 0.1 | The car is currently stopped |
| $0.1 \leq$ speed $< 2.0$ | The car is moving slowly |
| $2.0 \leq$ speed $< 4.5$ | The car is driving at a moderate speed |
| $4.5 \leq$ speed | The car is driving at a high speed |

**Table 10.** Throttle/brake discretion

| Throttle/brake | Text Prompt |
|---|---|
| $0.5 < brake$ | Braking hard |
| $0.05 < brake \leq 0.5$ | Braking |
| $0.8 < throttle$ | Accelerating fast |
| $0.3 < throttle \leq 0.8$ | Accelerating |
| $0.05 < throttle \leq 0.3$ | Accelerating gently |
| else | Idling |

**Table 11.** Steering discretion

| Steering | Text Prompt |
|---|---|
| $-0.05 < steering < 0.05$ | Going straight |
| $0.3 < steering \leq 1$ | Turning right sharply |
| $0.05 \leq steering \leq 0.3$ | Turning right |
| $steering < -0.3$ | Turning left sharply |
| $-1 \leq throttle \leq -0.3$ | Turning left |

## Declaration of competing interest

The authors declare that they have no known competing financial interests or personal relationships that could have appeared to influence the work reported in this paper.

## Acknowledgement

Funding for this research was provided by the Center for Connected and Automated Transportation under Grant No. 69A3552348305 of the U.S. Department of Transportation, Office of the Assistant Secretary for Research and Technology (OST-R), University Transportation Centers Program. General support was provided by the Center for Innovation in Control, Optimization, and Networks (ICON) and the Autonomous and Connected Systems (ACS) initiatives at Purdue University's College of Engineering. In addition, this research was supported by grants from NVIDIA and utilized NVIDIA RTX PRO 6000 Blackwell Max-Q Workstation Edition GPUs and A100 GPU-Hours on Saturn Cloud.

## Citations and References

Bai, S., Chen, K., Liu, X., Wang, J., Ge, W., Song, S., Dang, K., Wang, P., Wang, S., Tang, J., 2025. Qwen2. 5-vl technical report. arXiv preprint arXiv:2502.13923.

CARLA Team, n.d. CARLA Autonomous Driving Leaderboard [WWW Document]. URL https://leaderboard.carla.org/ (accessed 11.27.25a).

CARLA Team, n.d. CARLA Autonomous Driving Challenge [WWW Document]. URL https://leaderboard.carla.org/challenge/ (accessed 11.27.25b).

Chekroun, R., Toromanoff, M., Hornauer, S., Moutarde, F., 2023. Gri: General reinforced imitation and its application to vision-based autonomous driving. Robotics 12, 127.

Chen, D., Koltun, V., Krähenbühl, P., 2021. Learning to drive from a world on rails. Presented at the Proceedings of the IEEE/CVF International Conference on Computer Vision, pp. 15590–15599.

Chen, D., Krähenbühl, P., 2022. Learning from all vehicles. Presented at the Proceedings of the IEEE/CVF Conference on Computer Vision and Pattern Recognition, pp. 17222–17231.

Chitta, K., Prakash, A., Geiger, A., 2021. Neat: Neural attention fields for end-to-end autonomous driving. Presented at the Proceedings of the IEEE/CVF International Conference on Computer Vision, pp. 15793–15803.

Chitta, K., Prakash, A., Jaeger, B., Yu, Z., Renz, K., Geiger, A., 2022. Transfuser: Imitation with transformer-based sensor fusion for autonomous driving. IEEE transactions on pattern analysis and machine intelligence 45, 12878–12895.

Codevilla, F., Santana, E., López, A.M., Gaidon, A., 2019a. Exploring the limitations of behavior cloning for autonomous driving. Presented at the Proceedings of the IEEE/CVF international conference on computer vision, pp. 9329–9338.

Codevilla, F., Santana, E., López, A.M., Gaidon, A., 2019b. Exploring the limitations of behavior cloning for autonomous driving. Presented at the Proceedings of the IEEE/CVF international conference on computer vision, pp. 9329–9338.

Dosovitskiy, A., Ros, G., Codevilla, F., Lopez, A., Koltun, V., 2017. CARLA: An open urban driving simulator. Presented at the Conference on robot learning, PMLR, pp. 1–16.

Fujimoto, S., Gu, S.S., 2021. A minimalist approach to offline reinforcement learning. Advances in neural information processing systems 34, 20132–20145.

Fujimoto, S., Hoof, H., Meger, D., 2018. Addressing function approximation error in actor-critic methods. Presented at the International conference on machine learning, PMLR, pp. 1587–1596.

Guo, D., Yang, D., Zhang, H., Song, J., Zhang, R., Xu, R., Zhu, Q., Ma, S., Wang, P., Bi, X., 2025. Deepseek-r1: Incentivizing reasoning capability in llms via reinforcement learning. arXiv preprint arXiv:2501.12948.

Haarnoja, T., Zhou, A., Abbeel, P., Levine, S., 2018. Soft actor-critic: Off-policy maximum entropy deep reinforcement learning with a stochastic actor. Presented at the International conference on machine learning, Pmlr, pp. 1861–1870.

Hou, H., Zeng, P., Ma, F., Yu, F.R., 2025. Visualrwkv: Exploring recurrent neural networks for visual language models. Presented at the Proceedings of the 31st International Conference on Computational Linguistics, pp. 10423–10434.

Huang, S., Dossa, R.F.J., Ye, C., Braga, J., Chakraborty, D., Mehta, K., Araújo, J.G., 2022. Cleanrl: High-quality single-file implementations of deep reinforcement learning algorithms. Journal of Machine Learning Research 23, 1–18.

Huang, Z., Sheng, Z., Chen, S., 2025a. PE-RLHF: Reinforcement Learning with Human Feedback and physics knowledge for safe and trustworthy autonomous driving. Transportation Research Part C: Emerging Technologies 179, 105262.

Huang, Z., Sheng, Z., Ma, C., Chen, S., 2024a. Human as AI mentor: Enhanced human-in-the-loop reinforcement learning for safe and efficient autonomous driving. Communications in Transportation Research 4, 100127.

Huang, Z., Sheng, Z., Qu, Y., You, J., Chen, S., 2024b. VLM-RL: A Unified Vision Language Models and Reinforcement Learning Framework for Safe Autonomous Driving. arXiv preprint arXiv:2412.15544.

Huang, Z., Sheng, Z., Wan, Z., Qu, Y., Luo, Y., Wang, B., Li, P., Chen, Y.-J., Chen, J., Long, K., Meng, J., Leng, Y., Chen, S., 2025b. Sky-Drive: A distributed multiagent simulation platform for human−AI collaborative and socially aware future transportation. Journal of Intelligent and Connected Vehicles 8, 9210070. https://doi.org/10.26599/JICV.2026.9210070

Ilharco, G., Wortsman, M., Wightman, R., Gordon, C., Carlini, N., Taori, R., Dave, A., Shankar, V., Namkoong, H., Miller, J., Hajishirzi, H., Farhadi, A., Schmidt, L., 2021. OpenCLIP. https://doi.org/10.5281/zenodo.5143773

Jiang, K., Cai, X., Cui, Z., Li, A., Ren, Y., Yu, H., Yang, H., Fu, D., Wen, L., Cai, P., 2024. Koma: Knowledge-driven multi-agent framework for autonomous driving with large language models. IEEE Transactions on Intelligent Vehicles.

Kostrikov, I., Nair, A., Levine, S., 2021. Offline reinforcement learning with implicit q-learning. arXiv preprint arXiv:2110.06169.

Kumar, A., Zhou, A., Tucker, G., Levine, S., 2020. Conservative q-learning for offline reinforcement learning. Advances in neural information processing systems 33, 1179–1191.

Lake, B.M., Ullman, T.D., Tenenbaum, J.B., Gershman, S.J., 2017. Building machines that learn and think like people. Behavioral and brain sciences 40, e253.

Liang, E., Liaw, R., Nishihara, R., Moritz, P., Fox, R., Goldberg, K., Gonzalez, J., Jordan, M., Stoica, I., 2018. RLlib: Abstractions for distributed reinforcement learning. Presented at the International conference on machine learning, PMLR, pp. 3053–3062.

Long, K., Shi, H., Zhou, Y., Li, X., 2024. Physics Enhanced Residual Policy Learning (PERPL) for safety cruising in mixed traffic platooning under actuator and communication delay.

Nair, A., Gupta, A., Dalal, M., Levine, S., 2020. Awac: Accelerating online reinforcement learning with offline datasets. arXiv preprint arXiv:2006.09359.

Peng, B., Alcaide, E., Anthony, Q., Albalak, A., Arcadinho, S., Biderman, S., Cao, H., Cheng, X., Chung, M., Grella, M., 2023. Rwkv: Reinventing rnns for the transformer era. arXiv preprint arXiv:2305.13048.

Peng, Z., Li, Q., Liu, C., Zhou, B., 2022. Safe driving via expert guided policy optimization. Presented at the Conference on Robot Learning, PMLR, pp. 1554–1563.

Peng, Z.M., Mo, W., Duan, C., Li, Q., Zhou, B., 2024. Learning from active human involvement through proxy value propagation. Advances in neural information processing systems 36.

Qu, Y., Huang, Z., Sheng, Z., Chen, J., Chen, S., Labi, S., 2025a. VL-SAFE: Vision-Language Guided Safety-Aware Reinforcement Learning with World Models for Autonomous Driving. arXiv preprint arXiv:2505.16377.

Qu, Y., Xu, Z., Huang, Z., Sheng, Z., Chen, S., Chen, T., 2025b. MetaSSC: Enhancing 3D semantic scene completion for autonomous driving through meta-learning and long-sequence modeling. Communications in Transportation Research 5, 100184.

Radford, A., Kim, J.W., Hallacy, C., Ramesh, A., Goh, G., Agarwal, S., Sastry, G., Askell, A., Mishkin, P., Clark, J., 2021. Learning transferable visual models from natural language supervision. Presented at the International conference on machine learning, PmLR, pp. 8748–8763.

Raffin, A., Hill, A., Gleave, A., Kanervisto, A., Ernestus, M., Dormann, N., 2021. Stable-baselines3: Reliable reinforcement learning implementations. Journal of machine learning research 22, 1–8.

Seno, T., Imai, M., 2022. d3rlpy: An offline deep reinforcement learning library. Journal of Machine Learning Research 23, 1–20.

Shao, H., Wang, L., Chen, R., Li, H., Liu, Y., 2023. Safety-enhanced autonomous driving using interpretable sensor fusion transformer. Presented at the Conference on Robot Learning, PMLR, pp. 726–737.

Sheng, Z., Huang, Z., Chen, S., 2024. Traffic expertise meets residual RL: Knowledge-informed model-based residual reinforcement learning for CAV trajectory control. Communications in Transportation Research 4, 100142.

Sheng, Z., Huang, Z., Qu, Y., Chen, J., Luo, Y., Chen, Y.-J., Leng, Y., Chen, S., 2025a. Safeplug: Empowering multimodal llms with pixel-level insight and temporal grounding for traffic accident understanding. arXiv preprint arXiv:2508.06763.

Sheng, Z., Huang, Z., Qu, Y., Leng, Y., Bhavanam, S., Chen, S., 2025b. CurricuVLM: Towards Safe Autonomous Driving via Personalized Safety-Critical Curriculum Learning with Vision-Language Models. Transportation Research Part C: Emerging Technologies (accepted, in-press).

Sheng, Z., Huang, Z., Qu, Y., Leng, Y., Chen, S., 2025c. Talk2traffic: Interactive and editable traffic scenario generation for autonomous driving with multimodal large language model. Presented at the Proceedings of the Computer Vision and Pattern Recognition Conference, pp. 3788–3797.

Sutton, R.S., Barto, A.G., 1998. Reinforcement learning: An introduction. MIT press Cambridge.

Toromanoff, M., Wirbel, E., Moutarde, F., 2020. End-to-end model-free reinforcement learning for urban driving using implicit affordances. Presented at the Proceedings of the IEEE/CVF conference on computer vision and pattern recognition, pp. 7153–7162.

Toromanoff, M., Wirbel, E., Moutarde, F., 2019. Is deep reinforcement learning really superhuman on atari? leveling the playing field. arXiv preprint arXiv:1908.04683.

Wang, S., Wang, J., Wang, G., Zhang, B., Zhou, K., Wei, H., 2024. Open-vocabulary calibration for fine-tuned CLIP. arXiv preprint arXiv:2402.04655.

Wu, P., Jia, X., Chen, L., Yan, J., Li, H., Qiao, Y., 2022. Trajectory-guided control prediction for end-to-end autonomous driving: A simple yet strong baseline. Advances in Neural Information Processing Systems 35, 6119–6132.

Xu, Z., Zhang, Y., Xie, E., Zhao, Z., Guo, Y., Wong, K.-Y.K., Li, Z., Zhao, H., 2024. Drivegpt4: Interpretable end-to-end autonomous driving via large language model. IEEE Robotics and Automation Letters.

Yarats, D., Fergus, R., Lazaric, A., Pinto, L., 2021. Mastering visual continuous control: Improved data-augmented reinforcement learning. arXiv preprint arXiv:2107.09645.

Zhang, Z., Liniger, A., Dai, D., Yu, F., Van Gool, L., 2021. End-to-end urban driving by imitating a reinforcement learning coach. Presented at the Proceedings of the IEEE/CVF international conference on computer vision, pp. 15222–15232.

Zhang, Z., Tang, S., Zhang, Y., Fu, T., Wang, Y., Liu, Y., Wang, D., Shao, J., Wang, L., Lu, H., 2024. AD-H: Autonomous Driving with Hierarchical Agents. arXiv preprint arXiv:2406.03474.

Zheng, Y., Zhang, R., Zhang, J., Ye, Y., Luo, Z., Feng, Z., Ma, Y., 2024. LlamaFactory: Unified Efficient Fine-Tuning of 100+ Language Models, in: Proceedings of the 62nd Annual Meeting of the Association for Computational Linguistics (Volume 3: System Demonstrations). Association for Computational Linguistics, Bangkok, Thailand.

Zhu, J., Wang, W., Chen, Z., Liu, Z., Ye, S., Gu, L., Tian, H., Duan, Y., Su, W., Shao, J., 2025. Internvl3: Exploring advanced training and test-time recipes for open-source multimodal models. arXiv preprint arXiv:2504.10479.